\documentclass[10pt,twocolumn,letterpaper]{article}

\usepackage[pagenumbers]{cvpr} 

\usepackage{algorithm}
\usepackage{algorithmic}

\usepackage{amsmath}
\usepackage{booktabs}

\usepackage{multirow}
\usepackage{flushend}

\definecolor{cvprblue}{rgb}{0.21,0.49,0.74}
\usepackage[pagebackref,breaklinks,colorlinks,allcolors=cvprblue]{hyperref}

\def\paperID{****} 
\def\confName{CVPR}
\def\confYear{2026}

\title{From Sequential to Spatial: Reordering Autoregression \\  for Efficient Visual Generation}

\author{\hspace{-0.4cm}Siyang Wang$^{1}$ ~ Hanting Li$^{2}$ ~ Wei Li$^{2}$  ~ Jie Hu$^{2}$ ~ Xinghao Chen$^{2}$ ~ Feng Zhao$^{1}$\\ 
	$^{1}$University of Science and Technology of China\\
	$^{2}$Huawei Noah's Ark Lab\\
	{\tt\small \hspace{0cm}\{lihanting3,wei.lee,hujie23,xinghao\}@huawei.com}\quad\quad \\
    {\tt \small siyangw@mail.ustc.edu.cn, fzhao956@ustc.edu.cn}
}

\begin{document}
\twocolumn[{
    \renewcommand\twocolumn[1][]{#1}
    \maketitle
    \begin{center}
        \centering\captionsetup{type=figure}
        \includegraphics[width=0.88\linewidth]{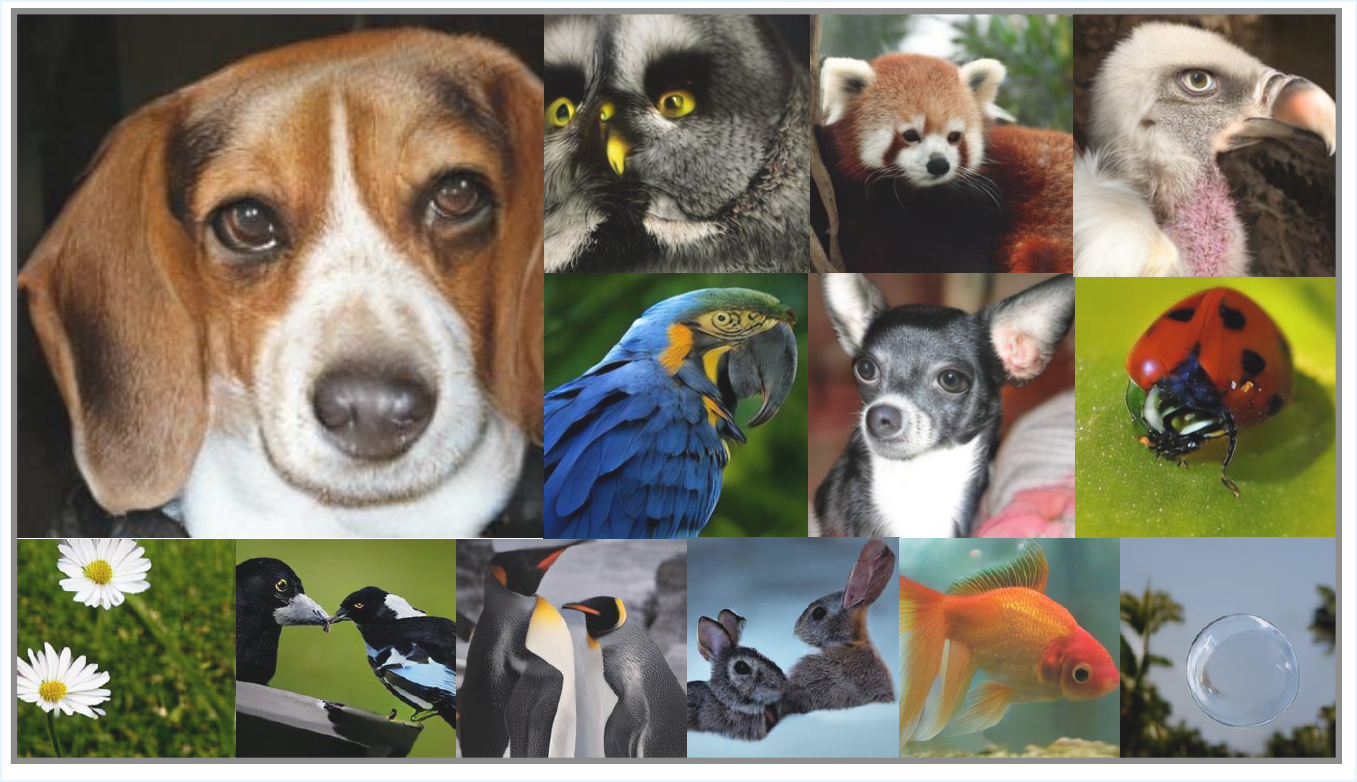}
        \captionof{figure}{Generated samples from RadAR-XL. We show samples at different aspect ratios (top) and zero-shot image editing results (bottom) including class-conditional editing and out-painting . Please zoom in for details.} 
        \label{fig1}
    \end{center}
}]

\begin{abstract} 
Inspired by the remarkable success of autoregressive models in language modeling, this paradigm has been widely adopted in visual generation. However, the sequential token-by-token decoding mechanism inherent in traditional autoregressive models leads to low inference efficiency.In this paper, we propose RadAR, an efficient and parallelizable framework designed to accelerate autoregressive visual generation while preserving its representational capacity. Our approach is motivated by the observation that visual tokens exhibit strong local dependencies and spatial correlations with their neighbors—a property not fully exploited in standard raster-scan decoding orders. Specifically, we organize the generation process around a radial topology: an initial token is selected as the starting point, and all other tokens are systematically grouped into multiple concentric rings according to their spatial distances from this center. Generation then proceeds in a ring-wise manner, from inner to outer regions, enabling the parallel prediction of all tokens within the same ring. This design not only preserves the structural locality and spatial coherence of visual scenes but also substantially increases parallelizability.
Furthermore, to address the risk of inconsistent predictions arising from simultaneous token generation with limited context, we introduce a nested attention mechanism. This mechanism dynamically refines implausible outputs during the forward pass, thereby mitigating error accumulation and preventing model collapse.
By integrating radial parallel prediction with dynamic output correction, RadAR reduces the number of inference steps from 256 to only 13, achieving up to a 5.6× speedup on the ImageNet dataset and significantly improving generation efficiency. 
\end{abstract}

\section{Introduction}
\label{sec:intro}

\begin{figure*}[ht]
\centering 
\includegraphics[width=0.9\linewidth]{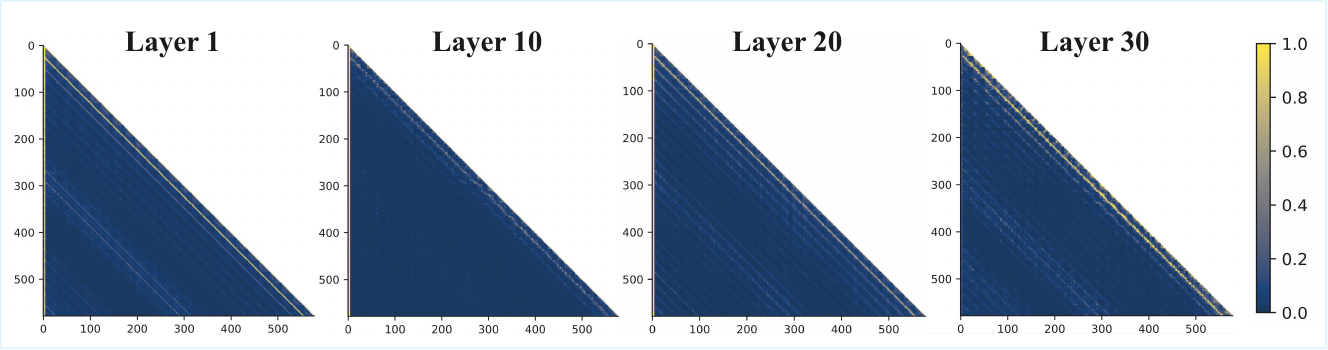}
\caption{The distribution of attention scores across visual tokens in different layers of the LlamaGen-XL~\cite{sun2024autoregressive} . Slash lines indicate that significant attention scores are allocated to tokens at fixed intervals, corresponding to tokens in the same column of previous rows.
} 
\label{fig2}
\end{figure*}

Autoregressive (AR) models demonstrate remarkable performance in language modeling~\cite{brown2020language,yang2025qwen3,dubey2024llama,liu2024deepseek} and exhibit considerable potential for the development of artificial general intelligence. To extend the success of AR models from language to visual domain, researchers\cite{esser2021taming,ramesh2021zero} begin to explore the application of AR models in image generation. Pioneering works such as VQVAE-2~\cite{razavi2019generating} and Parti~\cite{yu2022scaling} establish the fundamental feasibility of autoregressive visual generation by discretizing images into sequences of visual tokens, with subsequent research efforts~\cite{xie2024show,team2024chameleon,sun2024autoregressive} progressively pushing the performance boundaries. These models not only achieve visual fidelity comparable to or surpassing contemporary state-of-the-art diffusion models but, more significantly, provide architectural foundations for unified multimodal understanding and generation. This progress enables the development of integrated models capable of processing text, images, and video within a unified framework.
However, these performance improvements come with considerable computational demands. In conventional autoregressive approaches for visual generation, two-dimensional images are typically flattened into one-dimensional sequences following raster-scan order and processed through sequential token prediction.
This sequential processing mechanism establishes a quadratic relationship between the number of required inference steps and the image resolution, imposing significant constraints on generation efficiency. Consequently, deploying autoregressive models in practical scenarios faces substantial challenges, with generation speed becoming a major bottleneck hindering their widespread application, especially in scenarios requiring high-throughput requirements.

To address these efficiency limitations, the research community develops various methods for accelerating sequence generation. Existing methods generally fall into two primary categories. The first category attempts to break the strict sequential constraints of conventional autoregressive generation by allowing parallel prediction of multiple image tokens in a single inference step. The representative work MaskGIT~\cite{chang2022maskgit} innovatively integrates masked modeling strategies with non-autoregressive generation, achieving significant acceleration by iteratively predicting and refining tokens at multiple masked positions. Similarly, PAR~\cite{wang2025parallelized} further explores the potential of parallel decoding within random positions or block-constrained regions. Nevertheless, these methods still face limitations in the trade-off between generation quality and efficiency. The second category seeks breakthroughs from the perspective of feature representation. Works represented by VAR~\cite{tian2024visual} introduce multi-scale token representations, achieving an order-of-magnitude reduction in generation steps. However, this multi-scale token representation fundamentally differs in architecture from the universal flat token representation (e.g., CLIP~\cite{radford2021learning, zhai2023sigmoid}, DINO~\cite{oquab2023dinov2, caron2021emerging}), preventing direct utilization of general visual representations learned from large-scale vision-language pairs. This compatibility issue severely restricts the interoperability of such methods with modern multimodal systems~\cite{zhao2025qlip,tong2025metamorph,wu2024vila}, impeding their broad application in complex systems requiring collaboration with existing perceptual backbones.

To systematically investigate more efficient visual decoding mechanisms, this paper conducts an in-depth analysis of the attention distribution patterns in autoregressive visual models. Our analysis reveals that visual content inherently exhibits strong locality. As shown in Fig.~\ref{fig2}, although LlamaGen~\cite{sun2024autoregressive} is forced to follow the raster-scan sequence order during training, its attention distribution strongly reflects the inherent two-dimensional spatial structure of visual content. Specifically, when processing the current token, LlamaGen allocates significant attention weights not only to temporally adjacent tokens in the sequence but also to tokens that are temporally distant yet spatially proximate in the original image structure.  This observation emphasizes the importance of modeling two-dimensional spatial relationships directly, beyond linear sequences. The conventional raster-scan order imposes a one-dimensional constraint that conflict with natural visual structure, forcing the model to learn to compensate for this structural mismatch during training, ultimately degrading both generation efficiency and semantic consistency.

Building on the above insights, we propose RadAR—a radial parallel autoregressive generative framework designed to maintain full representational capacity while achieving efficient inference. The framework establishes an initial token as the radial center and systematically partitions remaining tokens into concentric rings based on spatial distance. The generation process progresses ring by ring radially from the inside out, with all tokens within the same ring decoded in parallel. This design significantly enhances parallelism while preserving the locality and spatial coherence of visual structures. However, parallel generation introduces critical challenges: when multiple tokens are decoded simultaneously without sufficient contextual constraints, semantically inconsistent predictions are prone to occur. To address this, we further design a nested attention mechanism that operates at both inter-ring and intra-ring levels. By iteratively refining unreasonable outputs, this mechanism effectively suppresses error propagation and enhances generation robustness.

Experimental results demonstrate that by integrating radial parallel prediction with dynamic output correction, RadAR effectively addresses the speed limitations inherent in traditional autoregressive models. The proposed framework achieves a speedup of up to 5.6× on the ImageNet dataset, offering a scalable and high-performance solution for next-generation visual content synthesis.

\section{Related Work}
\label{sec:Relate}

\begin{figure*}[ht]
\centering 
\includegraphics[width=\linewidth]{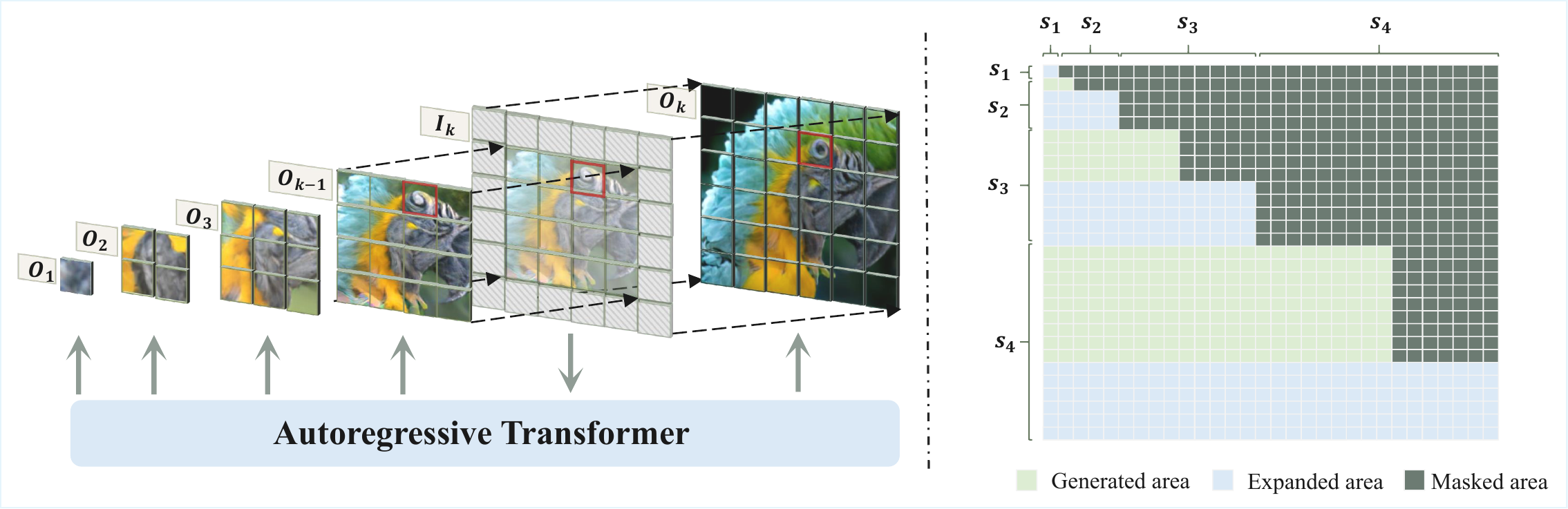}
\caption{Illustration of our radial-parallel autoregressive (RadAR) generation framework.  
(a) {\bf Model implementation.} RadAR takes the output $O_{k-1}$ from step $k-1$ and then uses learnable parameters to fill the range to be expanded outside $O_{k-1}$, forming the input $I_k$ for step $k$. For specific details on the extrapolation range, please refer to Algorithm 1.  
 (b) {\bf Nested attention mask.} For step $k$, green corresponds to the range where content has already been generated, and blue corresponds to the new unknown area expanded. The attention range is restricted to distinguish between the two purposes of error correction and content generation. Specifically, the newly expanded area in step $k$ accesses the context information of all tokens from previous $k-1$ steps and step $k$ to complete the generation purpose, while the generated area is restrict from accessing the expanded area.} \label{fig3}
\end{figure*}

\subsection{Autoregression in Language Modeling.}
The autoregressive modeling paradigm first emerges in the field of natural language processing (NLP)~\cite{vaswani2017attention,radford2019language,brown2020language,wan2023efficient,zhou2023survey}, with the core idea of generating sequences by predicting each element in a sequence based on the previous elements through conditional probabilities. However, this simple mechanism has demonstrated remarkable scalability across multiple dimensions. On one hand, it follows an observable and reproducible scaling law, where every leap in computing power, parameters, and data volume brings predictable performance gains. On the other hand, once the scale of parameters exceeds a critical threshold, the model suddenly emerges with zero-shot and few-shot generalization capabilities. Without fine-tuning for specific tasks, it can be applied to translation, question answering, reasoning, code generation, and even multimodal understanding and generation, thereby opening unprecedented prospects for the development of general artificial intelligence.

\subsection{Autoregression in Visual Generation.}
Autoregression has been explored in the field of visual generation for many years. Early pixel-based methods~\cite{salimans2017pixelcnn++,den2016conditional} applied raster-scan order to directly convert two-dimensional images into one-dimensional pixel sequences for pixel-by-pixel prediction. This paradigm is inefficient in generation and often produces suboptimal and blurry results due to the inherent redundancy of pixel information. The advent of token-based models marked a major breakthrough in visual autoregressive modeling, inspired by natural language processing (NLP)~\cite{brown2020language}. Unlike pixel-level models that directly handle raw visual data, this paradigm first compresses visual data into compact token sequences via discrete tokenizers \cite{esser2021taming,van2017neural,yu2023language}, and quantizes them to the nearest codes in a fixed-size codebook to form compact token sequences. It then autoregressively predicts these tokens in raster scan order~\cite{yu2022scaling,ramesh2021zero,lee2022autoregressive}.
However, the improvement in efficiency is still limited. Autoregressive visual generative models trained with a next-token prediction targets require $n^{2}$ sequential model forward passes to generate an image represented by $n\times n$ tokens, leading to a significant reduction in generation efficiency. 
This issue is particularly prominent in high-resolution image~\cite{cao2023efficient} or video generation~\cite{tulyakov2018mocogan,hong2022cogvideo}, hindering their widespread application in real-world applications. 
To accelerate sequence generation, researchers explore the parallel prediction methods.
Parallel prediction can generally be divided into two categories: raster-scan order and random-scan order. Among them, parallel prediction in raster-scan order targets a set of spatially regular tokens~\cite{wang2025parallelized,he2025neighboring}. Random-scan order is first proposed by MaskGIT~\cite{chang2022maskgit}, which employs a masked generation approach to generate next token set instead of next one token, substantially reducing the inference steps. However, when the number of inference steps is small, the generation quality of both types of efficient autoregressive methods becomes suboptimal. VAR is the first to break the limitation of inference steps on generation performance. By developing coarse-to-fine next-scale predictions, it transitions visual autoregressive modeling from the token level to the scale level, completing the generation process with only 10 steps. However, VAR relies on a multi-scale VQGAN~\cite{lee2022autoregressive,razavi2019generating}  that is tightly integrated with the generator and has been trained across a series of complex scales. Furthermore, the VAR generator is constrained by its complex and rigid multi-scale architecture. Modifying the output resolution or adjusting inference steps requires retraining both the tokenizer and generator, resulting in poor scalability. Additionally, its specialized multi-scale tokenizer creates compatibility barriers when integrating with large language models  for multimodal unification, significantly limiting flexibility.

\section{Method}
\label{sec:method}
{\bf Radial-Parallel Autoregressive Modeling.} Our proposed parallel autoregressive method is illustrated in Fig.~\ref{fig3}. Unlike the common raster order, the generation process is concentric, akin to progressively adding borders from the center to the surroundings, starting from scratch. At step $t$, the model receives the already generated rectangular area and adds a border of a certain thickness around it. This border is filled with a learnable prompt vector $ p \in {R^{c\times1\times1}}$, and then the generated area along with the prompt is processed as the input for the current step. This concentric stacking mechanism enables highly parallel expansion while maintaining strong spatial locality through radial dependencies. In the specific implementation, we first perform center cropping on the quantized latent feature $ f_q $ to obtain $ N $ token sets $\{g_1, g_2, \dots, g_N\}$ with incrementally larger spatial ranges, where $g_k = \{x_i\}_{i=0}^{n}$. The spatial range represented by these token sets increases progressively, with the final $ g_N $ matching the range of the original feature map $ R^{c \times h \times w} $, that is, $g_N = f_q$. Algorithm~\ref{alg:enc} provides a detailed description of the data preprocessing before autoregressive training. The autoregressive likelihood function is reformulated as:
\begin{equation}
p({g}_{1}, {g}_{2},\dots,{g}_{N})  = \displaystyle\prod_{i=1}^{N}p( {g}_{i}| {g}_{1}, {g}_{2},\dots,{g}_{i-1}).
\end{equation}
Crucially, our method enables dynamic context updating, allowing each generation step to optimize previously synthesized areas using the latest context information, thereby implementing an implicit error correction mechanism. This mitigates the accumulation of prediction errors common in parallel token generation, achieving an optimal balance between parallelization efficiency (allowing concurrent generation of spatially independent border segments) and maintenance of generation quality. As a result, compared to traditional parallel autoregressive methods, the number of decoding steps required is significantly reduced while avoiding their characteristic error propagation issues.

\begin{algorithm}[tb]
\caption{Radial-Parallel Autoregressive Encoding}
\label{alg:enc}
\textbf{Input}: raw image $im$, learnable vector $p$\\
\textbf{Parameter}: steps $N$, current patch nums ${\{P_i\}}_{i=1}^N$\\
\textbf{Output}: regression input token sets $R_{s}$ and regression target token sets $R_{gt}$
\begin{algorithmic}[1] 
\STATE $f = \mathcal{E}(im)$
\STATE $f_q = \mathcal{Q}(f)$
\STATE $i=1$
\WHILE{$i < N+1$}
\STATE $g_i = \text{crop}(f_q,P_i)$
\STATE $R_{gt} = \text{queue\_push}(R_{gt},g_i)$
\IF {$i < N$}
\STATE $S_{i+1} = \text{repeat}(p,P_{i+1})$     
\STATE $S_{i+1} = \text{center\_fill}(S_{i+1},g_i,P_i,P_{i+1})$  
\STATE $R_{s} = \text{queue\_push}(R_{s},S_{i+1})$
\ENDIF
\ENDWHILE
\end{algorithmic}
\end{algorithm}

\begin{figure*}[ht]
\centering 
\includegraphics[width=0.9\linewidth]{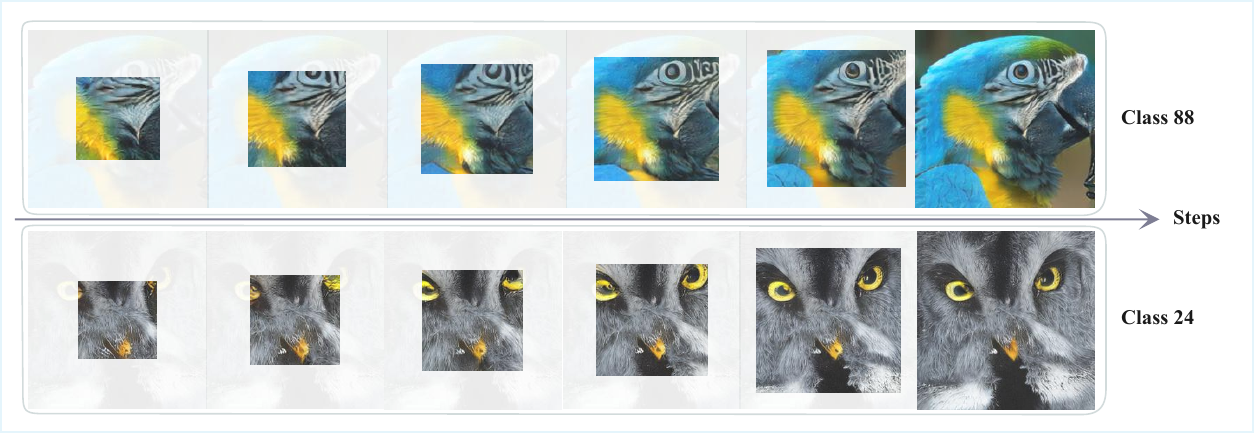}
\caption{Visualization of error correction. Earlier generation errors are gradually corrected by the subsequent generation process as the number of inference steps increases. Please zoom in
for details.
} \label{fig4}
\end{figure*}

{\bf Nested Attention Mechanism.}
To address the distinct objectives of the already generated area, which requires minor corrections for potential generation errors, and the newly expanded area, which necessitates generating content from scratch, we introduce the nested attention mechanism as illustrated in Fig.~\ref{fig3}. Specifically, causal attention is applied between steps to ensure the orderly progression of the autoregressive process, while bidirectional attention is employed within each step to maintain spatial consistency. Moreover, to prevent the error-prone newly expanded area from misleading the error correction decisions of the already generated area, causal attention is also maintained between the known and newly expanded areas within each step. During inference, KV-caching is used and no mask is needed. The experimental section compares the effects of using and not using this nested attention mechanism.

{\bf Tokenizer Post-training.} 
To mitigate the gap between the latent space sampled by the generator and the latent space seen during tokenizer training, we introduced a lightweight post-training phase. Deviating from the standard teacher-forcing technique, the input for the next autoregressive step is a composite of the output from the previous autoregressive step and the ground-truth, amalgamated in a specified ratio. The loss function is composed of two parts and is represented as:
\begin{equation}
{L}_{vqgan} = {L}_{recon}+{L}_{GAN}+{L}_{percp}+{L}_{vq},\nonumber,
\end{equation}
\begin{equation}
L = \frac{\lambda }{N}\displaystyle\sum_{n=1}^{N}cos({f}_{n}^{dec,l},\phi({f}_{n}^{DINO}) )+{L}_{vqgan}.
\end{equation}
where ${L}_{vqgan}$ includes ${L}_{recon}$ (the reconstruction loss), ${L}_{percp}$ (the perceptual loss), ${L}_{GAN}$ (the adversarial loss), and ${L}_{vq}$ (the codebook loss). To enable the tokenizer to generate latent representations with greater semantic consistency, we introduce the semantic regularization loss. Here, $N$ denotes the batch size, and $n$ represents the image index. An MLP is used to project the decoder feature ${f}_{n}^{dec,l}$ to match the channel dimension of the feature from a pre-trained visual encoder (here DINOv2-B). The semantic regularization loss aligns the features by computing the cosine similarity, thereby constraining the complexity of the latent space and preventing the tokenizer from learning overly complex latent dependencies.

\section{Experiments}
\label{sec:exp}

\begin{table*}[h]
\centering
\caption{Performance comparisons on class-conditional ImageNet 256×256 benchmark. Metrics are Frechet inception distance (FID), inception score (IS), precision and recall. “↓” or “↑” indicate lower or higher values are better. “Steps” denotes the number of model forward passes required to generate an image. MAR (step=10) is evaluated using code and pretrained weights from their official GitHub repository. \textbf{Bold} denotes the best, and \underline{underline} denotes the second.}
\label{tab:main}
\setlength{\tabcolsep}{3mm}
\begin{tabular}{clcccccc}
\toprule
\multicolumn{1}{c|}{\textbf{Type}}                                                           & \multicolumn{1}{c|}{\textbf{Model}} & \multicolumn{1}{c|}{\textbf{Params}}      & \textbf{Steps↓} & \textbf{FID↓}         & \textbf{IS↑}          & \textbf{Precision↑}   & \textbf{Recall↑}      \\ \midrule
\multicolumn{1}{c|}{\multirow{3}{*}{GAN}}                                                    & \multicolumn{1}{l|}{BigGAN~\cite{brocklarge}}         & \multicolumn{1}{c|}{112M}                 & 1              & 6.95                 & 224.5                & 0.89                 & 0.38                 \\
\multicolumn{1}{c|}{}                                                                        & \multicolumn{1}{l|}{GigaGAN~\cite{kang2023scaling}}        & \multicolumn{1}{c|}{569M}                  & 1             & 3.45                 & 225.5                & 0.84                 & 0.61                 \\
\multicolumn{1}{c|}{}                                                                        & \multicolumn{1}{l|}{StyleGAN-XL~\cite{sauer2022stylegan}}    & \multicolumn{1}{c|}{166M}                  & 1             & 2.30                  & 265.1                & 0.78                 & 0.53                 \\ \midrule

\multicolumn{1}{c|}{\multirow{4}{*}{Diff}}                                                   & \multicolumn{1}{l|}{ADM-G~\cite{dhariwal2021diffusion}}          & \multicolumn{1}{c|}{554M}                 & 250            & 4.59                 & 186.7                & 0.82                 & 0.52                 \\
\multicolumn{1}{c|}{}                                                                        & \multicolumn{1}{l|}{ADM~\cite{dhariwal2021diffusion}}            & \multicolumn{1}{c|}{554M}                 & 250            & 10.94                & 101.0                & 0.69                 & 0.63                 \\

\multicolumn{1}{c|}{}                                                                        & \multicolumn{1}{l|}{DiT-L/2~\cite{peebles2023scalable}}        & \multicolumn{1}{c|}{458M}                  & 250           & 5.02                 & 167.2                & 0.75                 & 0.57                 \\
\multicolumn{1}{c|}{}                                                                        & \multicolumn{1}{l|}{DiT-XL/2~\cite{peebles2023scalable}}       & \multicolumn{1}{c|}{675M}                 & 250            & 2.27                 & 278.2                & 0.83                 & 0.57                 \\ \midrule
\multicolumn{1}{c|}{\multirow{3}{*}{Mask}}                                                   & \multicolumn{1}{l|}{MaskGIT~\cite{chang2022maskgit}}        & \multicolumn{1}{c|}{227M}                 & 8              & 6.18                 & 182.1                & 0.8                  & 0.51                 \\
\multicolumn{1}{c|}{}                                                                        & \multicolumn{1}{l|}{Open-MAGVIT2~\cite{luo2024open}}   & \multicolumn{1}{c|}{343M}                 & 256            & 3.08                 & 258.3                & 0.85                 & 0.51                 \\
\multicolumn{1}{c|}{}                                                                        & \multicolumn{1}{l|}{MAR-H~\cite{li2024autoregressive}}          & \multicolumn{1}{c|}{943M}                 & 10             & 9.32                 & 207.4                & 0.71                 & 0.47                 \\ \midrule

\multicolumn{1}{c|}{\multirow{4}{*}{VAR}}                                                    & \multicolumn{1}{l|}{VAR-d16~\cite{tian2024visual}}        & \multicolumn{1}{c|}{310M}                 & 10             & 3.55                 & 280.4                & 0.84                 & 0.51                 \\
\multicolumn{1}{c|}{}                                                                        & \multicolumn{1}{l|}{VAR-d20~\cite{tian2024visual}}        & \multicolumn{1}{c|}{600M}                & 10            & 2.95                 & 302.6                & 0.83                 & 0.56                 \\ \cline{2-8} 
\multicolumn{1}{c|}{}                                                                        & \multicolumn{1}{l|}{FAR-B~\cite{yu2025frequency}}          & \multicolumn{1}{c|}{208M}                 & 10             & 4.26                 & 248.9                & 0.79                 & 0.51                 \\
\multicolumn{1}{c|}{}                                                                        & \multicolumn{1}{l|}{FAR-H~\cite{yu2025frequency}}          & \multicolumn{1}{c|}{812M}                 & 10             & 3.21                 & 300.6                & 0.81                 & 0.55                 \\ \midrule

\multicolumn{1}{c|}{\multirow{3}{*}{AR}} 

& \multicolumn{1}{l|}{LlamaGen-L~\cite{sun2024autoregressive}}     & \multicolumn{1}{c|}{343M}                 & 256            & 3.80                  & 248.3                & 0.83                 & 0.52                 \\
\multicolumn{1}{c|}{}                                                                        & \multicolumn{1}{l|}{LlamaGen-XL~\cite{sun2024autoregressive}}    & \multicolumn{1}{c|}{775M}                  & 256           & 3.39                 & 227.1                & 0.81                 & 0.54        \\
\multicolumn{1}{c|}{}                                                                        & \multicolumn{1}{l|}{LlamaGen-XXL~\cite{sun2024autoregressive}}   & \multicolumn{1}{c|}{1.4B}                 & 256            & 3.09        & 253.6                & 0.83                 & 0.53           \\
\midrule

\multicolumn{1}{c|}{\multirow{8}{*}{\begin{tabular}[c]{@{}c@{}}Parallelized\\ AR\end{tabular} }} 
& \multicolumn{1}{l|}{PAR-L-4×~\cite{wang2025parallelized}}       & \multicolumn{1}{c|}{343M}                 & 67             & 4.32                 & 189.4                & \textbf{0.87}        & 0.43                 \\
\multicolumn{1}{c|}{}                                                                        & \multicolumn{1}{l|}{PAR-XL-4×~\cite{wang2025parallelized}}      & \multicolumn{1}{c|}{775M}                 & 67             & 3.50                 & 234.4                & 0.84                 & 0.49                 \\
\cline{2-8} 

\multicolumn{1}{c|}{}                                                                        
& \multicolumn{1}{l|}{NAR-L~\cite{he2025neighboring}}       & \multicolumn{1}{c|}{372M}                 & 31             & 3.06                 & 263.9               & 0.81        & 0.53                 \\
\multicolumn{1}{c|}{}                                                                        & \multicolumn{1}{l|}{NAR-XL~\cite{he2025neighboring}}      & \multicolumn{1}{c|}{816M}                 & 31             & \textbf{2.70}                 & 277.5                & 0.81                 & \textbf{0.58}                 \\
\cline{2-8}

\multicolumn{1}{c|}{}                                                                        & \multicolumn{1}{l|}{RadAR}          & \multicolumn{1}{c|}{310M}                 & 10             & 3.50                 & \textbf{307.9}       & 0.81                 & 0.50                 \\
\multicolumn{1}{c|}{}                                                                        & \multicolumn{1}{l|}{RadAR}          & \multicolumn{1}{c|}{310M}                  & 13            & 3.36                 & 289.5                & 0.84                 & \underline{0.54}                \\
\multicolumn{1}{c|}{}                                                                        & \multicolumn{1}{l|}{RadAR}          & \multicolumn{1}{c|}{610M}                 & 10             & 3.21                 & 291.4                & 0.83                 & 0.52                 \\
\multicolumn{1}{c|}{}                                                                        & \multicolumn{1}{l|}{RadAR}          & \multicolumn{1}{c|}{610M}                 & 13             & \underline{2.97}           & \underline{ 303.8}          & \underline{ 0.85 }                & 0.53          \\ \bottomrule
                                                                                             &                                     & \multicolumn{1}{l}{} & \multicolumn{1}{l}{}                & \multicolumn{1}{l}{} & \multicolumn{1}{l}{} & \multicolumn{1}{l}{} & \multicolumn{1}{l}{} \\
                                                                                             &                                     & \multicolumn{1}{l}{} & \multicolumn{1}{l}{}                & \multicolumn{1}{l}{} & \multicolumn{1}{l}{} & \multicolumn{1}{l}{} & \multicolumn{1}{l}{} \\
                                                                                             &                                     & \multicolumn{1}{l}{} & \multicolumn{1}{l}{}                & \multicolumn{1}{l}{} & \multicolumn{1}{l}{} & \multicolumn{1}{l}{} & \multicolumn{1}{l}{} \\
                                                                                             &                                     & \multicolumn{1}{l}{} & \multicolumn{1}{l}{}                & \multicolumn{1}{l}{} & \multicolumn{1}{l}{} & \multicolumn{1}{l}{} & \multicolumn{1}{l}{} \\
                                                                                             &                                     & \multicolumn{1}{l}{} & \multicolumn{1}{l}{}                & \multicolumn{1}{l}{} & \multicolumn{1}{l}{} & \multicolumn{1}{l}{} & \multicolumn{1}{l}{}
\end{tabular}
\vspace{-6em}

\end{table*}

In this section, we first introduce the experimental setup, including datasets, metrics, baselines, and implementation details. Then, we compare RadAR with state-of-the-art methods, followed by ablation studies to analyze the effectiveness of each component and some zero-shot generation experiments to demonstrate the flexibility of RadAR. 

\subsection{Experimental Settings}
{\bf Implementation Details.} Our tokenizer is initialized with the pre-trained weights of llamaGen~\cite{sun2024autoregressive}, with the codebook size set to 8912. It downsamples the input image at a fixed ratio of 16×16. We mainly follow the VQVAE training recipe of llamaGen and fine-tune on ImageNet-1K~\cite{deng2009imagenet} with randomly sampled resolutions. RadAR is also trained on the ImageNet-1K~\cite{deng2009imagenet} 256×256 dataset. The training process employs the AdamW optimizer~\cite{adam2014method} with ${\beta }_{1}=0.9$, ${\beta}_{2}=0.95$, and a weight decay rate of 0.05. The learning rate is set to 1e-4, and the training epochs are uniformly set to 300 epochs.

{\bf Baselines.} We choose baseline methods from popular image generation models, including generative adversarial networks (GAN.) ~\cite{brocklarge,kang2023scaling,sauer2022stylegan}, diffusion models(Diff.)~\cite{dhariwal2021diffusion,radford2018improving}, masked-prediction models (Mask.) ~\cite{chang2022maskgit,luo2024open,li2024autoregressive},visual autoregressive models (VAR.)~\cite{tian2024visual,yu2025frequency} and autoregressive models (AR.)~\cite{sun2024autoregressive,wang2025parallelized,he2025neighboring}. These methods have achieved impressive results on various image generation tasks and serve as strong baselines for comparison.  Since the prediction of our method is token-based and aligns with the objectives of AR methods like LlamaGen~\cite{sun2024autoregressive}, we take it as a strong baseline and primarily compare our method with it.

{\bf Evaluation Metrics.} We adopt four metrics for quantitative evaluation: Frechet Inception Distance (FID)~\cite{heusel2017gans}, Inception Score (IS)~\cite{salimans2016improved}, Precision, and Recall~\cite{kynkaanniemi2019improved}. Following the standard protocols, we sample 50,000 images with trained models to evaluate.

\begin{figure*}[ht]
\centering 
\includegraphics[width=0.8\linewidth]{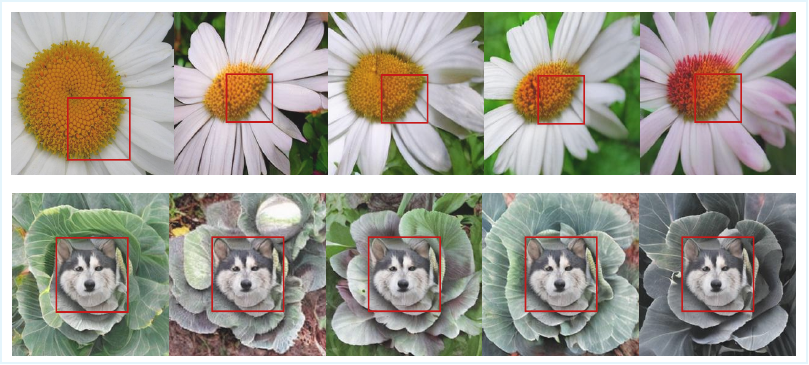}
\caption{Zero-shot evaluation in downstream tasks containing out-painting and class-conditional editing. The results show that RadAR can generalize to novel downstream tasks without special design and finetuning. Please zoom in for details.
} \label{fig5}
\end{figure*}

\subsection{Class-conditional Image Generation}
In this section, we evaluate the performance of RadAR on the ImageNet, with the results shown in Table~\ref{tab:main}. We provide two scales of RadAR, L and XL. Compared with state-of-the-art models, RadAR shows significant competitiveness in multiple key metrics, including inference steps, FID, IS, precision, and recall. Among AR methods, RadAR significantly reduced the number of inference steps required for image generation. Specifically, compared with the 256 steps required for the token-by-token generation process of LlamaGen and the 67 steps required for the parallel generation mechanism of PAR, RadAR reduces the number of inference steps to approximately 10. This reduction is not only quantitative but also accompanied by improved generation performance, making RadAR a leading candidate for efficient and high-quality image synthesis. Compared with the multi-scale VAR methods, although the FID and recall scores of RadAR are slightly lower than those of VAR, it achieves higher IS and Precision scores. This indicates that RadAR excels in generating images with higher perceptual quality and more accurate details, although it may sacrifice overall diversity and recall to some extent.

Since RadAR's prediction target is the ground-truth value, this design brings multiple benefits. On the one hand, RadAR can dynamically adjust the number of inference steps according to the desired generation effect and computational efficiency. In contrast, methods that predict residuals, such as VAR, need to retrain the tokenizer and autoregressive generator if the number of inference steps is to be changed. This adaptability highlights RadAR's strong generalization ability. On the other hand, this modeling approach is highly consistent with the requirements of multimodal vision-language models (VLMs). By directly predicting the ground-truth value, RadAR ensures that the generated images are not only perceptually realistic but also semantically consistent with the input prompts. This seamless integration with VLMs is an advantage that residual modeling methods do not possess. Overall, RadAR achieves a balanced and advanced level in terms of inference efficiency and generation quality, making it a strong competitor in the field of autoregressive visual generation.

\subsection{Ablation Study}
{\bf Component-wise Ablations.} In Table~\ref{tab:ab}, we conduct comprehensive ablation studies to investigate the effectiveness of our key designs on the ImageNet 256×256 dataset. Specifically, these include: 1) Nested Attention Mechanism (NAM): This attention mechanism disentangles the functions of error correction and ground-truth value prediction, thereby improving the FID score. 2) Random Dropout Sampling (RDS): This training sampling strategy not only enhances RadAR's ability to perceive arbitrary inference ranges but also increases the flexibility of the model's inference steps. 3) Random Noise Input (RNI): This technique increases the model's tolerant ability for deviations between generated results and ground-truth during inference, thereby improving the score of IS.
4) Tokenizer Post-training (TPT): This mechanism effectively mitigates the gap between the latent space sampled by the generator and the latent space, resulting in significant improvement of each evaluation metrics.

\begin{table}[]
\centering
\caption{Ablations on the effectiveness of the proposed techniques. Specific meanings of the abbreviations are in the ablation part. We report the results with RadAR-XL.}
\vspace{-0.5em}
\setlength{\tabcolsep}{1.8mm}
\begin{tabular}{cccc|cccc}
\toprule
\multicolumn{4}{c|}{\textbf{Ablations}}                                           & \multicolumn{4}{c}{\textbf{Metrics}} \\ \midrule
NAM        & RDS        & RNI  & TPT     & FID↓     & IS↑     & Pre↑     & Rec↑     \\ \midrule
$\times$            & $\times$            &$\times$    &$\times$         &5.70         &255.0        &0.81         & 0.41        \\
$\surd$            &$\times$             &$\times$      &$\times$       &3.80         &271.0        &0.80         &0.50         \\
$\surd$           & $\surd$           &$\times$      &$\times$       & 3.60        &295.4        &0.80         &0.52         \\
$\surd$           & $\surd$           & $\surd$     &$\times$      &3.55         &297.6        &0.80         &0.52         \\ 
$\surd$           & $\surd$           & $\surd$     &$\surd$      &3.31         &299.4        &0.82         &0.53         \\
\bottomrule
\end{tabular}
\label{tab:ab}
\end{table}

{\bf Visualization of Error Correction.} Autoregressive parallel generation improves the efficiency of visual generation but also raises the probability of token errors. This is one of the primary reasons why the performance of previous methods drops sharply when the autoregressive steps are highly compressed. Unlike traditional autoregressive methods, RadAR can adaptively detect and correct generation errors in the early stages of inference while generating new tokens. As shown in Fig.~\ref{fig4}, as the number of inference steps increases, some minor errors in the early stages are identified and corrected with the help of richer context information. For example, the parrot's feathers become clearer, and the shapes of the owl's eyes and beak become more normal.

{\bf Initial Token Position.} To evaluate the influence of initial token positions in the parallel generation, we conduct ablation studies comparing different starting location configurations, with quantitative results presented in Table~\ref{tab:pos}. The experiments demonstrate that initialization from the center position yields superior performance relative to edge-based initialization, which in turn surpasses corner-based initialization. This observed hierarchy suggests that the performance advantage is closely associated with the symmetrical propagation of contextual information during inference. Beyond quantitative metrics, the center-token initialization strategy also enables the most balanced high-parallelism computation, as radial expansion originating from the center facilitates uniform distribution of computational load across directional paths. These collective findings support the adoption of center-oriented initialization as the optimal configuration in the RadAR, effectively harmonizing generation quality with computational efficiency.

{\bf Image Generation Speed.} In Table~\ref{tab:speed}, we compare the generation speed of different autoregressive models with similar numbers of parameters on ImageNet 256 × 256. The results demonstrate that RadAR achieves a maximum speedup of approximately 5.6× relative to LlamaGen, while achieving generation efficiency comparable to VAR. More notably, RadAR maintains this competitive speed advantage without sacrificing architectural compatibility with mainstream visual token representations. This represents a notable practical advantage over VAR, which necessitates substantial modifications to the token representation architecture to attain similar efficiency improvements, consequently constraining its integration potential with existing multimodal systems. Collectively, these findings indicate that RadAR effectively mitigates the fundamental efficiency limitations of traditional autoregressive models while circumventing the compatibility challenges associated with alternative parallel generation methodologies. The framework establishes an optimal balance between generation speed, output quality, and system integration flexibility, positioning it as a viable solution for practical applications demanding high-throughput visual content synthesis.

\begin{table}[]
\centering
\caption{Ablations on the  initial token position (RadAR-XL). }
\vspace{-0.5em}
\setlength{\tabcolsep}{5.2mm}
\begin{tabular}{lccc}
\toprule
\textbf{Position}   & center & edge & corner \\ \midrule
\textbf{FID↓}                                     & 3.31             & 3.45                                                                 & 3.51              \\
\textbf{IS↑}                                     & 299.4             & 293.5                                                                 & 288.4              \\ \bottomrule 
\end{tabular}
\label{tab:pos}
\end{table}

\begin{table}[]
\centering
\caption{Class-conditional image \textbf{ generation  speed}.}
\setlength{\tabcolsep}{1.8mm}
\vspace{-0.5em}
\begin{tabular}{lcc|cc}
\toprule
\textbf{Model}         & \textbf{Params}       & \textbf{Steps} & \textbf{\begin{tabular}[c]{@{}c@{}}Throughput↑\\ (img/s)\end{tabular}} & \textbf{IS↑} \\ \midrule
VAR                    & 310M                  & 10             & 112.4                                                                 & 280.4       \\
LlamaGen             & 343M                  & 256            & 20.3                                                                  & 248.3       \\
PAR                    & 343M                  & 67             & 41.7                                                                  & 189.4       \\
NAR                    & 372M                  & 31             & 65.2                                                                  & 263.9       \\
RadAR       &310M   & 10             & 113.6                                                                 & 307.9
                       \\ \bottomrule
\end{tabular}
\label{tab:speed}
\end{table}

\subsection{Zero-shot Task Generalization}

{\bf Image Out-painting.} We place the ground-truth tokens in the unmasked area and let the model generate the tokens in the masked area. Fig.~\ref{fig5} (top) shows a set of results from RadAR-XL. Without any fine-tuning, RadAR can successfully complete the outpainting task, with natural transitions at the boundary junctions, demonstrating RadAR's good generalization ability.

{\bf Class-conditional Image Editing.} Furthermore, RadAR is applied to the class-conditional image editing task. Similar to the outpainting task, we place the ground-truth tokens in the unmasked area. The difference is that we use specific class labels as conditional to guide the generation of tokens in the masked area. The visualization results are shown in Fig.~\ref{fig5} (bottom), where RadAR can naturally blend images from different categories, once again demonstrating its powerful generation capability.

\begin{figure}[!ht]
\centering 
{\includegraphics[width=0.9\linewidth]{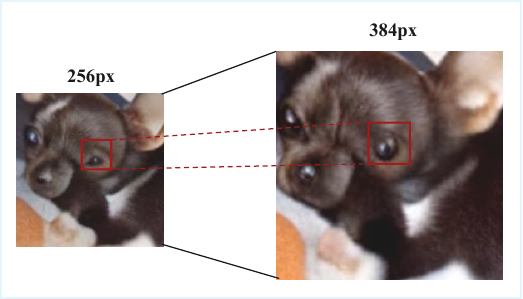}}
\caption{Visualization of Generation at high resolution. RadAR-XL (trained at 256px) is used. Zoom in for details.} 
\label{fig7}
\end{figure}

{\bf Generate Images at high Resolution.} In Fig.~\ref{fig7}, we visualize the result of zero-shot generation at high-resolution. We take low-resolution images as the input, and RadAR can generate high quality high-resolution images. Compared to VAR~\cite{tian2024visual}, which requires retraining the tokenizer and Autoregressive generator completely, our RadAR, despite being trained only at the 256×256 resolution, can accomplish this task without 
 fine-tuning, thanks to its inherent capability of both generation and refinement.
 
\vspace{-1.2em}
\vspace{-1.2em}

\section{Conclusion}
\label{sec:Conclusion}
In this paper, we introduce RadAR, which aims to address the limitations of traditional autoregressive models in terms of efficiency and flexibility. RadAR employs a unique inside-out, radial-parallel decoding strategy that ensures spatial consistency. Moreover, RadAR can dynamically correct existing generation errors during the inference process, thereby significantly reducing the number of inference steps required for image generation while maintaining high visual quality. Additionally, RadAR demonstrates strong generalization ability by successfully completing zero-shot image-to-image generation tasks and generating images with various aspect ratios, showing its competitiveness in the field of autoregressive visual generation.

{
    \small
    \bibliographystyle{ieeenat_fullname}
    \bibliography{main}
}

\end{document}